\documentclass[sigconf]{acmart}

\AtBeginDocument{%
  \providecommand\BibTeX{{%
    \normalfont B\kern-0.5em{\scshape i\kern-0.25em b}\kern-0.8em\TeX}}}

\copyrightyear{2021}
\acmYear{2021}
\setcopyright{acmcopyright}\acmConference[OASIS '21]{Proceedings of the 2021 Workshop on Open Challenges in Online Social Networks}{August 30, 2021}{Virtual Event, Ireland}
\acmBooktitle{Proceedings of the 2021 Workshop on Open Challenges in Online Social Networks (OASIS '21), August 30, 2021, Virtual Event, Ireland}
\acmPrice{15.00}
\acmDOI{10.1145/3472720.3483620}
\acmISBN{978-1-4503-8632-6/21/08}

\usepackage{times}
\usepackage{helvet}
\usepackage{courier}
\usepackage{graphicx} 
\usepackage{fancyhdr}
\usepackage{etoolbox}
\usepackage[export]{adjustbox}
\usepackage{float}
\usepackage{amsmath}
\usepackage{multirow}

\usepackage{color}

\usepackage{booktabs,chemformula}
\usepackage{hyperref}
\usepackage{latexsym}

\usepackage{paralist}
\usepackage[T1]{fontenc}
\usepackage[utf8]{inputenc}
\usepackage{microtype}
\usepackage{soul}

\newcommand\BibTeX{B\textsc{ib}\TeX}
\usepackage{subfig}

\begin{document}
\fancyhead{}
\title{Opinions are Made to be Changed:\\Temporally Adaptive Stance Classification} 

\author{Rabab Alkhalifa}
\orcid{0000-0002-2875-5400}
\affiliation{%
  \institution{Queen Mary University of London}
  \streetaddress{Mild End Road}
  \city{London}
  \country{UK}}
\email{r.a.a.alkhalifa@qmul.ac.uk}

\author{Elena Kochkina}
\orcid{0000-0003-0691-3647}
\affiliation{%
  \institution{Queen Mary University of London}
  \streetaddress{Mild End Road}
  \city{London}
  \country{UK}}
\email{e.kochkina@qmul.ac.uk}
  
\author{Arkaitz Zubiaga}
\orcid{0000-0003-4583-3623}
\affiliation{%
  \institution{Queen Mary University of London}
  \streetaddress{Mild End Road}
  \city{London}
  \country{UK}}
\email{a.zubiaga@qmul.ac.uk}

\renewcommand{\shortauthors}{Alkhalifa et al.}

\begin{abstract}

Given the rapidly evolving nature of social media and people's views, word usage changes over time. Consequently, the performance of a classifier trained on old textual data can drop dramatically when tested on newer data. While research in stance classification has advanced in recent years, no effort has been invested in making these classifiers have persistent performance over time. To study this phenomenon we introduce two novel large-scale, longitudinal stance datasets. We then evaluate the performance persistence of stance classifiers over time and demonstrate how it decays as the temporal gap between training and testing data increases. We propose a novel approach to mitigate this performance drop, which is based on temporal adaptation of the word embeddings used for training the stance classifier. This enables us to make use of readily available unlabelled data from the current time period instead of expensive annotation efforts. We propose and compare several approaches to embedding adaptation and find that the Incremental Temporal Alignment (ITA) model leads to the best results in reducing performance drop over time.

\end{abstract}



\maketitle



\section{Introduction}



Word meanings drift over time, with new words emerging, words adopting new senses and the frequency of word usage varying. Vocabulary and usage patterns in social media  evolve rapidly \cite{Hamoodat_VC2020}, and people's views change over time \cite{kelman1961processes}. This can have an impact on stance classification in social media as the data used for training may not generalise well to future data with different patterns. Previous research has either assumed that a classifier trained on static, temporally-restricted data would suffice to track public opinion over time \cite{deng2013tracking}, or focused on short time periods, analysing stance on trending topics such as Brexit, death penalty or climate change~\cite{Simaki2017a,Mohammad2016b}. Our work contributes to research in stance classification by focusing on the impact of a hitherto overlooked aspect: time.


A recent study by Florio et al.~\cite{florio2020time} demonstrated that social media hate speech detection models do not perform well on newer data when simply trained on older data. Despite highlighting the existence of this problem, their work did not propose any solutions to the problem.
Here we show that this problem is not exclusive to hate speech detection and that it also impacts the performance of social media stance classifiers \cite{alkhalifa2020qmul}. We collect two longitudinal stance detection datasets that we use for the classifier performance evaluation over time (Section 4). In our experiments we reproduce a real world scenario in which training data remains unchanged while new testing data is generated over years. Our findings indicate that a regular stance classifier can drop up to 18\% in relative performance in only five years (Sections 6, 7).  
We then propose novel methodology that makes a social media stance classifier more robust when applied to data that is temporally distant from the training data, which would in turn enable improved tracking of public opinion.

While one can choose the costly option of annotating new stance data regularly to re-train a classifier, here we investigate the scenario where one needs to make the most of the originally labelled data, e.g. due to limited resources. Hence we propose to use temporally adapted word embeddings, to re-train the classifier on the unchanged training data. This approach adapts the model to the vocabulary changes that happened over time while making use of readily available unlabelled data.
We compare two types of approaches to update the word embeddigns: (1) incrementally updating the same embedding model with new unlabelled data over time and (2) creating a temporally contextualised embedding for the testing year by incrementally aligning a new embedding with preceding embedding models over time. We find that the second approach is more successful at mitigating the performance drop over time. We can obtain improved performance with a substantially reduced performance drop of up to 5\%.


\section{Related Work}

\textbf{Stance classification.} There is a body of work on target-specific stance classification \cite{Mohammad2017a,Somasundaran2009} aiming to determine a user's supporting or opposing view towards a target. This research generally focused on a specific target \cite{Kucuk2019} and investigated as a static problem, looking at datasets that cover limited periods of time without paying attention at the impact of time. Others have looked at the problem of dealing with new targets on cross-target stance classification \cite{xu2018cross}, i.e. having training data associated with a particular target (e.g. Donald Trump), exploring the possibility of adapting the classifier to new targets (e.g. Joe Biden). 
Our research differs from this body of work in that we aim to (1) investigate the impact of time in stance classification for a particular target, and to (2) propose a model that makes this longitudinal tracking of stance more robust to changes in opinion \cite{sayeed2013opinion,Graells2020} and language \cite{croft2013evolution}, and hence more stable in performance. A line of research in stance identification has looked at the evolving nature of stance in rumour conversations \cite{lukasik2016hawkes,zubiaga2018discourse}, however this work focuses on stance exchanges in temporally brief conversations, rather than longitudinal persistence of models.

\textbf{Temporal persistence of classifiers.} Previous research has shown that classifiers trained on old data can drop in performance when tested on new data, as is the case with Amazon reviews \cite{Lukes2019} or hate speech detection \cite{florio2020time}. Works by \citet{Rocha2008} and \citet{Nishida2012} also find that the temporal gap between training and test data has a big impact on the performance of a classification model. Work by \citet{Nishida2012} proposed a multinomial naive Bayes classifier which switches between two probability estimates based on changes in word frequencies. Similarly, \citet{Preotiuc-Pietro2013} model periodic distributions of words over time for the hashtag prediction task using Gaussian Processes. Previous work however assumes that new labelled data is available over time, and therefore the classifier can be adapted using new labelled instances progressively; in our work we assume the realistic scenario where we have a labelled dataset pertaining to a period of time, and access to new labelled data for subsequent periods of time is not affordable. We tackle this problem by using word embeddings, which have been used in previous work for capturing semantic shift \cite{Zhang2016,Tan,kim2014temporal} (i.e. determining whether a word has changed its meaning over time), but there is a dearth of research exploring the use of embeddings to achieve persistence of classifiers. 


\section{Task Definition}


We define the stance classification task as identifying the attitude of the author of a post towards a certain topic as either supporting or opposing. Our task in this paper is to maximise the performance of a stance classifier when tested on the new data, which is several years apart from the training data, i.e. make the classifier persistent in time. 
Our proposed approach is based on adapting the word embeddings used to train the classifier, and thus we refer to is as adaptive stance classification. 
To study this we use (1)\textbf{ a longitudinal, unlabelled dataset} $D$, divided into $T$ equally sized temporal slices where $\mathrm{D}=\left\{D_{1}, D_{2}, \ldots, D_{T}\right\}$, and (2) \textbf{a longitudinal, labelled dataset} of annotated stance tweets representing temporal utterances from a particular domain (e.g., gender equality, healthcare) with a corresponding set of binary stance labels $s\in\{support, oppose\}$ spanning $T$ years, $Y = \{y_1, ..., y_T\}$, where $y_t$ is a set of tweets from year $t$. 
We use the unlabelled data to generate a sequence of temporal embeddings  $\mathrm{X}=\left\{X_{1}, X_{2}, \ldots, X_{T}\right\}$, where each  $X_{t}, t \in [1,T]$ contains vector representations of words generated using the temporal slice $D_{t}$ representing the ground truth of temporal representation at time $t$.
We assess the persistence of a classifier performance by training on data from one of the years $y_i$, where $i \in \{1, ..., t-1\}$ and testing it on each of the subsequent years $y_j$, where$j \in \{t+1, ..., T\}$. Our objective is to update the representation so as to adapt it to the vocabulary change and to maximise persistence in stance classification for any pair $y_i$ and $y_j$.

\section{Data} 

We use two types of datasets for our work: (1) labelled datasets, to assess stance detection models, and (2) larger unlabelled datasets, for building temporal word embedding models. Both types of datasets cover the same time period, enabling experiments on stance over time (labelled) by incrementally adapting word embeddings (unlabelled).

\begin{table}[t]
\centering
\begin{adjustbox}{max width=0.8\textwidth}
\begin{tabular}{|l|l|l|l|l|l|}
\hline
& \multicolumn{2}{|c|}{\textbf{Source}} & \textbf{Target} & \multicolumn{2}{|c|}{\textbf{Labels}} \\
\hline \hline
  &  \textbf{Train.} & \textbf{Eval.} & \textbf{Test.} & \textbf{\% Support} & \textbf{\% Oppose} \\ 	\hline\hline
\textbf{GE.} &  35100 & 3900 & 9000 & 76.9\% & 23.1\% \\
\hline \hline
\textbf{H.}  &  22500 & 2500 & 5040 & 53.6\% & 46.4\% \\
\hline
\end{tabular}
\end{adjustbox}
\caption{ Dataset statistics for Gender Equality (GE) and Healthcare (H) per year. Tweet counts for source and target for each year. Originally collected tweets were downsampled to the same number of supporting and opposing tweets per year.}
\label{tab:stance-dataset-corpora}
\end{table}

\begin{figure}[h]
    \centering
     \includegraphics[width=1\columnwidth]{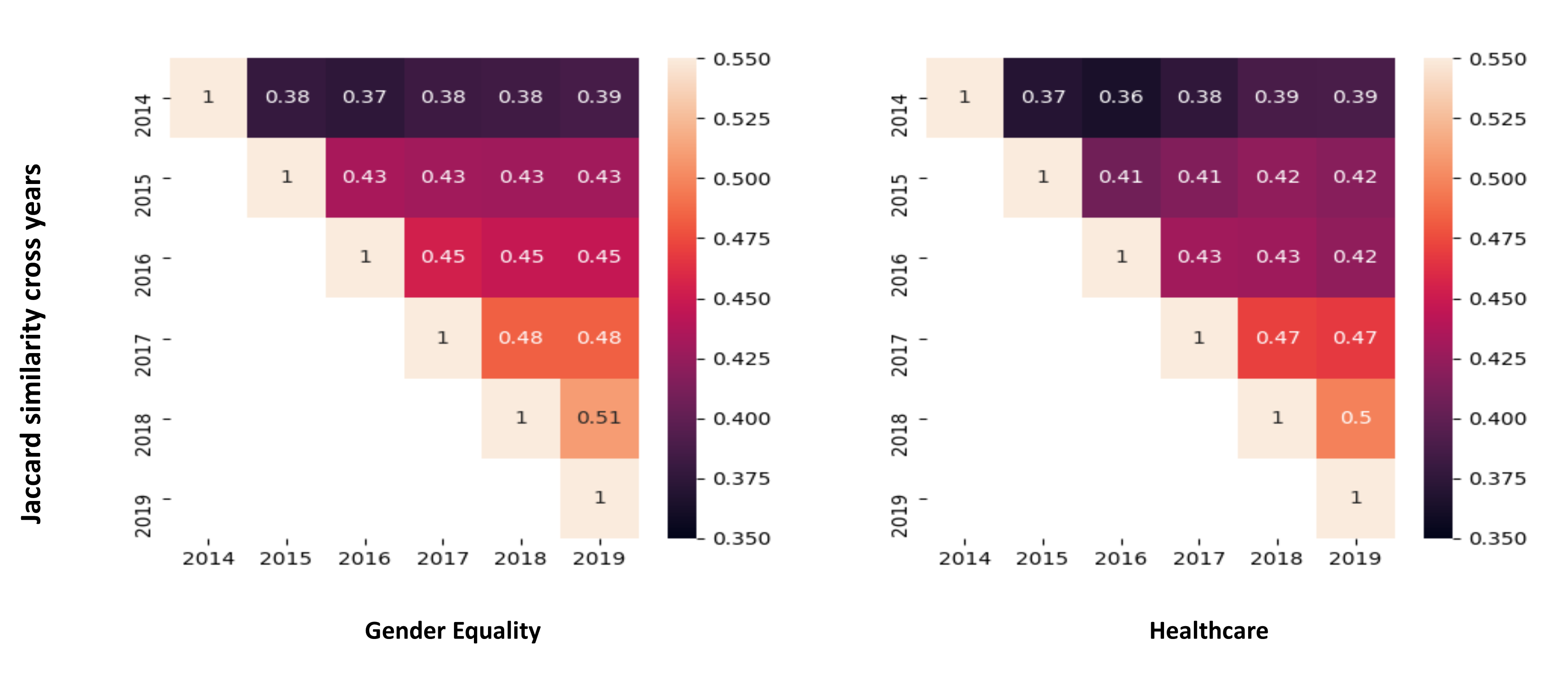}
    \caption{ Jaccard similarity between vocabularies for test sets of our annotated data collections.}
    \label{fig:jaccard}
\end{figure}



\begin{figure*}[h]

    \centering
    \includegraphics[height=0.27\textheight,width=0.7\textwidth]{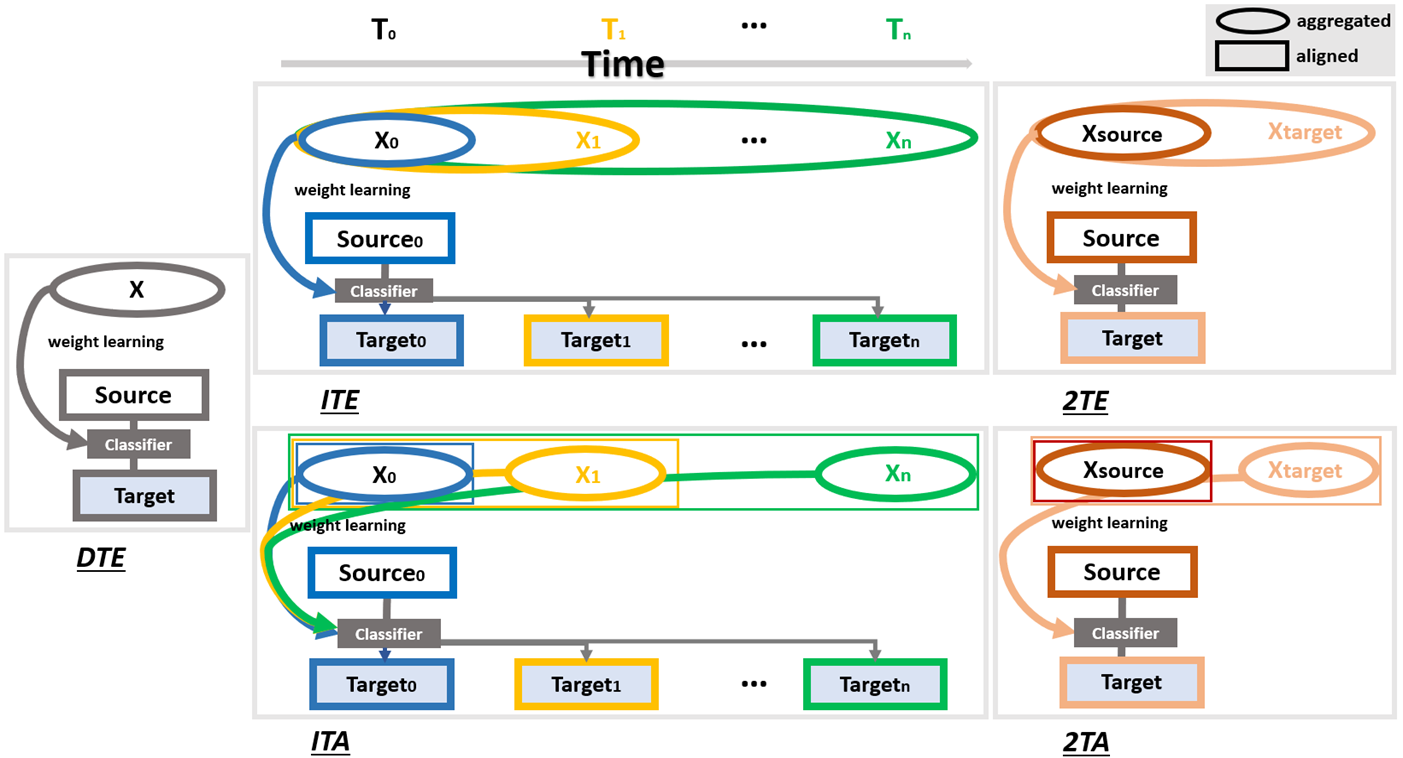}
    \caption{ Our adaptive stance classification models. DTE represents the baseline, static classifier. The other four are temporally evolving classifiers: 2TA and 2TE only consider source and target periods to model the evolution, whereas ITA and ITE consider all the periods between the source and the target.}
    \label{fig:experiment}
\end{figure*}

\subsection{Labelled Datasets}

Due to the lack of large-scale temporally annotated datasets for stance classification, we collected new datasets.
To enable collection of labelled datasets with sufficient data for each of the years under study, we opted for retrieving distantly supervised datasets, in this case for a six-year time period from 2014 to 2019. The data collection is based on predominantly supporting or opposing hashtags. 

Distant supervision became popular for collection of social media datasets labelled for sentiment \cite{go2009twitter} and has more recently been extended to other tasks, including stance classification \cite{Kumar2018,Mohammad2017a}. Distant supervision consists of defining a set of keywords (e.g. hashtags) which serve as a proxy to data labels, subsequently removing these keywords from the resulting dataset and leaving the rest of the text of the posts. We collected two Twitter stance datasets by using hashtags\footnote{ \url{https://github.com/OpinionChange2021/opinion_are_made_to_be_changed.git}} spanning the same time period (2014-2019): (1)  with hashtags supporting and opposing \textbf{Gender Equality}, involving issues such as feminism and gender pay gap, and (2) with hashtags supporting and opposing \textbf{Healthcare}, involving issues such as dieting and medical care. To assess the quality of the distantly supervised labels, we manually inspected a subset of 225 random tweets from the resulting datasets. We observed that only 11\% of the instances are noisy, i.e. opposite stance. This is in line with previous work on distant supervision (cf. \cite{purver2012experimenting}).

We randomly selected a stratified sample from each year, which is split into train, evaluation and test data. Table \ref{tab:stance-dataset-corpora} shows the per-year statistics of the resulting datasets.

To measure the temporal evolution of the datasets, we compute the Jaccard similarities between the vocabulary observed for each year. Figure \ref{fig:jaccard} shows the pairwise Jaccard similarity scores for the two datasets. We can observe that these similarity scores consistently decrease as the distance between the years increases, indicating an increasing variation of vocabulary over time.

\begin{table}[htp]
    \centering
    \begin{adjustbox}{max width=.5\textwidth}
    \begin{tabular}{|l | c | c | c|}
    \hline
     & \textbf{Embedding} & \textbf{Learning Strategy} & \textbf{Data} \\
     \hline
     \hline
     \textbf{DTE} & Discrete & None & S \\
     \hline
     \hline
     \textbf{ITE} & Incremental & Model Update & SPT \\
     \textbf{2TE} & Incremental & Model Update & ST \\
     \hline
     \hline
     \textbf{ITA} & Incremental & Diachronic Alignment & SPT \\
     \textbf{2TA} & Incremental & Diachronic Alignment & ST \\
     \hline
    \end{tabular}
     \end{adjustbox}
    \caption{ Temporal embedding models. Data sources: source year (S), target year (T), and preceding years (P).}
    \label{tab:methods}
   
\end{table}

\subsection{Unlabelled Datasets}

We also collected larger domain-specific Twitter datasets linked to the same two topics and using the same hashtags, however disregarding labels to avoid supervision when training the word embedding models. This resulted in 578K and 343K aggregated tweets for \textbf{Gender Equality} and  \textbf{Healthcare}, respectively.

\begin{figure*}[htp]%
    \centering
    \subfloat[\centering  ]{{\includegraphics[width=.4\textwidth]{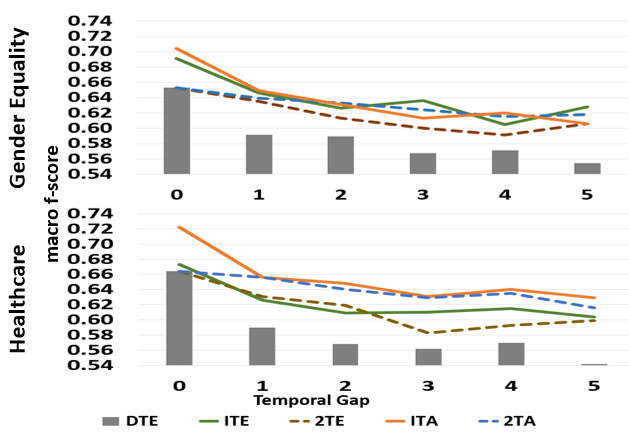} }}%
    \qquad
    \subfloat[\centering ]{{\includegraphics[width=.4\textwidth]{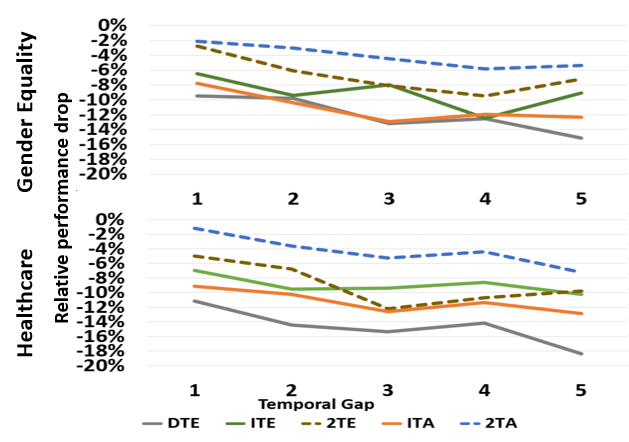} }}%
    \caption{ Performance (a) of temporal embeddings by temporal gap between training and test data (0-5 years) and (b) Relative performance drop. Flat line indicates consistent temporal performance.}%
    \label{fig:results}%
\end{figure*}

\newcommand{\STAB}[1]{\begin{tabular}{@{}c@{}}#1\end{tabular}}

\begin{table*}[h]
\centering
\begin{adjustbox}{max width=0.8\textwidth}
\begin{tabular}{|l|l|l|l|l|l|l|l|}
\hline
\multicolumn{2}{| c |}{\textbf{Time gap}} & \multicolumn{1}{c |}{\textbf{0}}     & \multicolumn{1}{c |}{\textbf{1}}     & \multicolumn{1}{c |}{\textbf{2}}     & \multicolumn{1}{c |}{\textbf{3}}     & \multicolumn{1}{c |}{\textbf{4}}     & \multicolumn{1}{c |}{\textbf{5}}     \\
\hline
\hline

\multirow{5}{*}{\STAB{\rotatebox[origin=c]{90}{\textbf{Gen. Equality}}}} & \textbf{DTE} &		0.653 &		0.591 (-9.5\%) &		0.589 (-9.8\%) &		0.567 (-13.2\%) &		0.571 (-12.6\%) &		0.554 (-15.2\%) \\
 & \textbf{ITE} &		0.691 &		0.646 (-6.5\%) &		0.626 (-9.4\%) &		\textbf{0.636} (-8.0\%) &	0.605 (-12.4\%) &		\textbf{0.628} (-9.1\%) \\
 & \textbf{2\-TE} &		0.653 &		0.635 (-2.8\%) &		0.613 (-6.1\%) &		0.600 (-8.1\%) &		0.591 (-9.5\%) &		0.606 (-7.2\%)\\
 & \textbf{ITA} &		\textbf{0.704} &		\textbf{0.649} (-7.8\%) &		0.631 (-10.4\%) &		0.613 (-12.9\%) &		\textbf{0.620} (-11.9\%) &		0.617 (-12.4\%) \\
 & \textbf{2\-TA}  &	0.653 &		0.639 (-2.1\%) &		\textbf{0.633} (-3.1\%) &		0.624 (-4.4\%) &		0.615 (-5.8\%) &		0.618 (-5.4\%)\\
\hline
\hline

\multirow{5}{*}{\STAB{\rotatebox[origin=c]{90}{\textbf{Healthcare}}}} & \textbf{DTE} &		0.664 &		0.590 (-11.1\%) &		0.568 (-14.5\%)	 &	0.562 (-15.4\%) &		0.570 (-14.2\%) &		0.542 (-18.4\%) \\
 & \textbf{ITE} &		0.673	 &	0.626 (-7.0\%) &		0.609 (-9.5\%) &		0.610 (-9.4\%) &		0.615 (-8.6\%) &		0.604 (-10.2\%) \\
 & \textbf{2\-TE} &	0.664	& 0.631 (-5.0\%)	& 0.619 (-6.8\%)	& 0.583 (-12.2\%)	& 0.593 (-10.7\%)	& 0.599 (-9.8\%) \\
 & \textbf{ITA} &		\textbf{0.722}	 &	\textbf{0.656} (-9.1\%) &		\textbf{0.648} (-10.2\%)	 &	\textbf{0.631} (-12.6\%)	 &	\textbf{0.640} (-11.4\%) &		\textbf{0.629} (-12.9\%) \\
 & \textbf{2\-TA} &		0.664 &		\textbf{0.656} (-1.2\%) &		0.640 (-3.6\%) &		0.629 (-5.3\%) &		0.635 (-4.4\%) &		0.616 (-7.2\%) \\

\hline 
\end{tabular}
\end{adjustbox}
\caption{ Experiment results by temporal gap between training and test data (0-5 years). Reported values in brackets indicate relative performance drops with respect to same-year (temporal gap of 0) experiments for the same method.}
\label{tab:rq2}
\end{table*}


\section{Methods for Incorporating temporal knowledge into word embeddings}

We assess the potential of word embeddings \cite{mikolov2013distributed} to aid classifiers to have a temporally persistent performance, and propose novel methods to further their temporal persistence (see Figure \ref{fig:experiment}). We use the CBOW model \cite{Mikolov2013a}, which outperformed skip-gram \cite{Mikolov2013a} for linguistic change detection \cite{kulkarni2015statistically}. We control for other variables (e.g. prediction models, label distributions) by keeping them stable across experiments.


\textit{\textbf{Method 1. Discrete Temporal Embedding (DTE),}} a baseline method that lacks awareness of the temporal evolution. DTE learns CBOW word vector representations given a collection of tweets pertaining to a particular time frame as input. For example, where our classifier needs to train from data pertaining to year $y_1$ and test it on $y_2$, a DTE embedding is generated from the unlabelled data pertaining to $y_1$. We can formally represent it as follows: Discrete Temporal Embedding (DTE) are the embeddings $X$ generated using temporal slice $D_s$ where $s$ represents the time frame of the source set.

In this work we propose four models to incorporate knowledge over time by leveraging unlabelled data. 

\textit{\textbf{Method 2. Incremental Temporal Embedding (ITE).}} New embeddings $X$ are trained using the unlabelled data incrementally aggregated from all years preceding and including the target year, i.e. $D_p$, where $p \in [2014,t]$ and $t$ is the target year. Then the stance classifier is retrained using the labelled data from the source year represented using the new up-to-date embeddings $X$.

\textit{\textbf{Method 3. Source-Target Temporal Embedding (2TE).}} New embeddings $X$ are generated using the unlabelled data aggregated from the source $D_s$ and the target $D_t$ years  only,  while ignoring all years in between. These embeddings are then used to represent source year training data for the stance classifier. 

While ITE and 2TE incorporate temporal knowledge, they do not explicitly handle other phenomena such as semantic shift of vocabulary \cite{kim2014temporal}, which we anticipate may lead to performance limitations. To address this, we propose alternative methods that perform temporal word alignment. Our proposed solution comes from using a \textit{compass}~\cite{Valerio2019Compass} method for temporal alignment. With \textit{compass} each temporal embedding becomes temporally contextualised to the testing year semantic-meaning. With this method, we assume words contextual usage fluctuates over time as in social associations creating subtle meaning drift. For example, the word \textit{`Clinton'} shifted from being related to administration to presidential context over time \cite{Valerio2019Compass}. The \textit{compass} aligns the embeddings of different temporal years using pivot non-shifting vocabularies. It constructs a dynamic temporal context embedding matrix that changes over time, allowing the context embedding to be more time relevant. These settings allow natural selection of vocabularies in terms of temporal contextual words of the target year, and a time-aware representation of the target year in general. We show that this approach is more useful in some cases than model update as the model trained considering the semantic meaning of the target year without additional contexts.





\textit{\textbf{Method 4. Incremental Temporal Alignment (ITA).}} $X$ is incrementally aligned using \textit{compass} from all preceding $D_p$, where $p \in [2014,t]$. 

\textit{\textbf{Method 5. Source-Target Temporal Alignment (2TA).}} $X$ performs temporal alignment using \textit{compass} of $D_s$ and $D_t$.

We summarise all five models in Table \ref{tab:methods} and Figure \ref{fig:experiment}, which enable us to test the impact of three different parameters: (i) the use of discrete vs incremental embedding models, (ii) the use of different learning strategies (none, model update, diachronic alignment), and (iii) the use of different data sources for building embedding models (source, target and preceding years).



\section{Experimental Setup} 


To control for the impact of model choice \cite{Lukes2019,Rocha2008}, we consistently use a Convolutional Neural Network (CNN) model  with 32 filter and 5-gram region sizes. Our $5$-gram kernels encompass a Rectified Linear Unit (ReLU) activation function, and a max-pooling operation. We use a softmax activation function, the Adam optimiser with the learning rate fixed at $2e^{-5}$ and 10 epochs. Sentence length for vector representations is also fixed in all experiments to $32$. While keeping the CNN model intact, our aim is to assess the effectiveness of the proposed embedding-based representations for the task.

We experiment with all 21 possible combinations from 2014 to 2019 of $y_{source}$ and $y_{target}$ for training and testing. In each case, we are interested in the temporal gap between train and test data, measured in number of years. For brevity and clarity, we report the mean average performance of models with the same gap (e.g. 4-year gap performance averages 2014-2018 and 2015-2019). In each case, we report the macro-averaged F1 score, as well as the Relative Performance Drop (RPD) to measure the sharpness of the drop in our model, defined as:

\[
            \operatorname{RPD}= \frac{f_{{score}_{t_j}} - f_{{score}_{t_0}}}{f_{{score}_{t_0}}}
            \]

Where $t_0$ represents  performance when temporal gap is 0; $t_j$ represents performance when temporal gap is one of 1-5.

\section{Results and Discussion}

Table \ref{tab:rq2} and Figure \ref{fig:results} show the results of our experiments. Results are aggregated by temporal gap. We observe that the best results are obtained for same-period experiments (i.e. temporal gap of 0) and a decrease in performance as the temporal gap increases, i.e. confirming our hypothesis that model persistence drops as training data gets older. Furthermore, the performance drops (percentage, shown in brackets) indicate that the drop has an upwards tendency for larger temporal gaps, demonstrating that the older the training model, the less accurate the model becomes when dealing with new data. Temporal dynamics in the stance datasets can indeed lead to deterioration in model persistence.

When we look at the methods separately, we observe that ITA achieves an overall best performance. This is especially true for the healthcare dataset, where ITA is the best method for all temporal gaps under consideration; for the gender equality dataset, ITA is the best method for small temporal gaps (0-1), and while it achieves competitive performance for larger gaps, 2TA and ITE occasionally achieve better performance.

We observe interesting trends when we look at performance scores and performance drops in conjunction. The use of the baseline DTE, solely relying on source-year embeddings, leads to the lowest performance and also the largest performance drops. This reinforces that embeddings from a particular time period gradually become less useful for subsequent periods, more so when the target period is more distant in time. Among the four proposed methods, ITA yields the best same-period performance, however it is also the method experiencing the highest performance drop for larger temporal gaps; this demonstrates ITA's competitive performance for shorter temporal gaps, but its performance on longer temporal gaps is more uncertain. For methods relying on source and target years, 2TE and 2TA, we observe a modest performance for same-period experiments (equivalent to DTE), which however experience a substantially smaller performance drop for larger temporal gaps. While their performance is not as good for small temporal gaps, they show a good capacity to persist better over time for larger temporal gaps. A look at longer temporal gaps, beyond the 5-year gap considered in our experiments, would be an interesting avenue for future work, e.g. to assess the capacity of 2TA and 2TE to persist further.



 




In addition, our experiments help us assess the impact of three parameters (see Table \ref{tab:methods}):

\textbf{Embedding type:} we show that the use of an incremental aggregation of embeddings (ITE, 2TE, ITA, 2TA) improves over the use of discrete embeddings (DTE). This is consistent across datasets and temporal gaps.

\textbf{Learning strategy:} our results indicate that the best learning strategy is the use of diachronic alignment (ITA, 2TA), in our case tested using \textit{compass}. With a few exceptions, we observe that these methods generally outperform methods that perform incremental model updates (2TE, ITE), and consistently outperform the lack of a learning strategy by relying on discrete embeddings (DTE).

\textbf{Data source:} the worst performance is for the embedding method solely using the source year (DTE), a baseline method that one would naturally use with static classifiers. Other methods considering additional years lead to improved performance. We observe two main patterns: (1) use of all years preceding the target year (ITE, ITA) lead to improved performance over the use of source and target years (2TE, 2TA), however with a larger performance drop for longer temporal gaps, and (2) use of source and target years only leads to lower performance in short temporal gaps, however with a substantially lower performance drop showing a promising trend towards achieving model persistence.

\section{Conclusion}

Our work demonstrates the substantial impact of temporal evolution on stance classification in social media, with performance drops of up to 18\% in relative macro-F1 scores in only five years. We investigate temporal adaptation of word embeddings used to thrain the classifier to mitigate this drop in performance, showing that incrementally aligning embedding data for all years (ITA) leads to the best performance. However, we also find that consideration of only source and target years in the alignment leads to the smallest performance drop with promising trends towards longer term persistence.

Furthering this research, we aim to investigate the extent to which different factors (e.g. opinion change, social media use) impact performance drop, as well as to explore the potential of using few-shot learning to quantify the benefits of labelling small amounts of target data.

\section{Acknowledgments}

First author would like to thank Ahmad Alkhalifah, Prof. Maria Liakata and Prof. Massimo Poesio for their constructive feedback throughout initial stage of this research work. This research utilised Queen Mary's Apocrita HPC facility, supported by QMUL Research-IT. 




\bibliographystyle{ACM-Reference-Format}
\bibliography{base.bib}

\end{document}